\documentclass[conference]{IEEEtran}
\IEEEoverridecommandlockouts

\RequirePackage[absolute]{textpos}
\DeclareRobustCommand*{\copyrightnote}{%
  \begin{textblock}{85}(17.5,259.5)
      \scriptsize{\noindent \copyright 2021 IEEE. Personal use of this material is permitted. Permission from IEEE must be obtained for all other uses, in any current or future media, including reprinting/republishing this material for advertising or promotional purposes, creating new collective works, for resale or redistribution to servers or lists, or reuse of any copyrighted
component of this work in other works.}%
  \end{textblock}%
    }

\usepackage{cite}
\usepackage{amsmath,amssymb,amsfonts}
 \usepackage{hyperref}

\usepackage[pdftex]{graphicx}
\usepackage{textcomp}
\usepackage{xcolor}
\usepackage{multirow}
\usepackage{float}
\usepackage{subcaption}
\usepackage{algorithm}
\usepackage{algpseudocode}
\usepackage[noend]{algcompatible}
\algnewcommand\algorithmicreturn{\textbf{return} }
\algnewcommand\RETURN{\State \algorithmicreturn}%
\algnewcommand\algorithmicto{\textbf{to} }
\algnewcommand\TO{\algorithmicto}%

\usepackage{varwidth}

\usepackage{booktabs}  
\usepackage{tablefootnote}
\def\BibTeX{{\rm B\kern-.05em{\sc i\kern-.025em b}\kern-.08em
  T\kern-.1667em\lower.7ex\hbox{E}\kern-.125emX}}

\title{Training Classifiers that are Universally Robust to All Label Noise Levels
\thanks{This research is supported by the National Research Foundation, Singapore under its AI Singapore Programme (AISG Award No: AISG-RP-2019-015), and under its NRFF Programme (Award No: NRFFAI1-2019-0005).}
}

\author{\IEEEauthorblockN{Jingyi Xu}
\IEEEauthorblockA{\textit{Singapore University} \\ \textit{of Technology and Design}\\
Singapore \\
jingyi\_xu@mymail.sutd.edu.sg}
\and
\IEEEauthorblockN{Tony Q. S. Quek,~\IEEEmembership{Fellow,~IEEE}}
\IEEEauthorblockA{\textit{Singapore University} \\ \textit{of Technology and Design}\\
Singapore \\
tonyquek@sutd.edu.sg}
\and
\IEEEauthorblockN{Kai Fong Ernest Chong}
\IEEEauthorblockA{\textit{Singapore University} \\ \textit{of Technology and Design}\\
Singapore \\
ernest\_chong@sutd.edu.sg}
}

\begin{document}
\maketitle
\copyrightnote


\begin{abstract}
For classification tasks, deep neural networks are prone to overfitting in the presence of label noise. Although existing methods are able to alleviate this problem at low noise levels, they encounter significant performance reduction at high noise levels, or even at medium noise levels when the label noise is asymmetric. To train classifiers that are universally robust to all noise levels, and that are not sensitive to any variation in the noise model, we propose a distillation-based framework that incorporates a new subcategory of Positive-Unlabeled learning. In particular, we shall assume that a small subset of any given noisy dataset is known to have correct labels, which we treat as ``positive", while the remaining noisy subset is treated as ``unlabeled". Our framework consists of the following two components: (1) We shall generate, via iterative updates, an augmented clean subset with additional reliable ``positive" samples filtered from ``unlabeled" samples; (2) We shall train a teacher model on this larger augmented clean set. With the guidance of the teacher model, we then train a student model on the whole dataset. Experiments were conducted on the CIFAR-10 dataset with synthetic label noise at multiple noise levels for both symmetric and asymmetric noise. The results show that our framework generally outperforms at medium to high noise levels. We also evaluated our framework on Clothing1M, a real-world noisy dataset, and we achieved 2.94\% improvement in accuracy over existing state-of-the-art methods.
\end{abstract}

\section{Introduction}
For classification tasks, deep neural networks (DNNs) are able to achieve zero training error when trained on datasets with label noise, even in the extreme scenario of totally randomized labels \cite{zhang2016understanding}.
This poses a challenge when training on real-world datasets. Manual data labeling 
is both inefficient and expensive, while automated annotation methods inherently introduce label noise. 
In either case, we typically have access to a small clean subset of the dataset.
For such datasets with label noise, how then do we train DNNs that generalize well, regardless of the actual (unknown) noise levels?

Label noise in real-world datasets is inevitably asymmetric. This asymmetry naturally arises because the performance of any data annotation method that is based on the output labels of a prediction model, is necessarily both class-dependent and instance-dependent \cite{Cheng2018:SurveyAutoImageAnnotation}. Although there are numerous existing methods for tackling label noise~\cite{reed2014training, ghosh2017robust, patrini2017making, hendrycks2018using, tanaka2018joint, zhang2018generalized, arazo2019unsupervised, shu2019meta, wang2019symmetric, xu2019l_dmi, zhang2019metacleaner,li2020dividemix,ma2020normalized}, all of which perform well in the idealized case of symmetric label noise (provided the noise level is not too high), these existing methods are not as robust to asymmetric label noise, and they exhibit a sharp performance drop at medium-to-high noise levels.

Despite much progress on tackling label noise, there are two seemingly opposing challenges that have not been tackled jointly: How do we train a model that is (i) robust to all noise levels, and (ii) whose performance at any given noise level is not sensitive to any variation in the noise model?

To solve both challenges via a unified approach, we propose a distillation-based framework that incorporates a new method of Positive-Unlabeled (PU) learning. 
In general, we have a trade-off between increasing accuracies across all noise levels for a given noise model, and increasing accuracies across all noise models for a given noise level. 
Our motivation to use PU learning is based on the observation that samples of any dataset with label noise can naturally be partitioned into ``correct" and ``incorrect" classes.
Clean data is correct by definition, while the remaining noisy dataset contains both correct and incorrect instances.
This is precisely the scenario of PU learning, where instances in the noisy dataset are treated as ``unlabeled".

Our framework comprises two components: clean data augmentation and knowledge distillation.
Starting with our given clean subset, we initially treat all instances with noisy labels (henceforth called ``noisy samples'') as being unlabeled, and we gradually augment the original clean subset, iteratively, via PU learning; those (unlabeled) noisy samples inserted into our augmented clean set would be assigned new labels. 
Crucially, this new label assignment does not require the originally given labels of the noisy samples. 
In other words, we \emph{do not require any assumptions on the underlying noise model.} Hence, this clean data augmentation component is automatically robust to all noise models. 
As for our distillation component, we train a teacher model solely on the augmented clean set, thus we are able to suppress the influence of label noise when training the student model, especially at high noise levels.

Our major contributions are summarized as follows:
\begin{itemize}
\item We propose a versatile distillation-based framework for tackling label noise.
In contrast to existing work, our framework is robust to all noise levels, and is not sensitive to noise model variation. To the best of our knowledge, this is the first ever solution that overcomes both major challenges we described earlier.

\item We introduce a new type of PU learning.
We then use this new technique to design a clean data augmentation algorithm, which also allows us to correct noisy samples with high confidence.
The effectiveness of our augmentation algorithm is demonstrated by the high precision scores of reliably clean samples extracted from the validation set.

\item In experiments on CIFAR-10 \cite{krizhevsky2009learning} with asymmetric semantic label noise, our proposed framework outperformed outperforms state-of-the-art (SOTA) methods at noise levels $50\%$--$90\%$. When evaluated on the real-world noisy dataset Clothing1M~\cite{xiao2015learning}, we achieved a new SOTA accuracy of 77.70\% (2.94\% higher than previous SOTA).
\end{itemize}

\section{Related Work}
\label{sec: related work}
\subsection{Existing Approaches for Tackling Label Noise}
\noindent\textbf{Data cleaning methods.} 
These are methods applied to identified noisy labels, and they include 
label correction~\cite{tanaka2018joint}, 
sample re-weighting~\cite{shu2019meta, jiang2018mentornet, zhang2019metacleaner}, 
label distribution estimation~\cite{yi2019probabilistic}, re-sampling of a relabeled dataset~\cite{wu2018light}, and treating ambiguous samples as unlabeled data and applying a semi-supervised method~\cite{Ding2018AST}.
The efficacy of such methods depends heavily on the identification of noisy labels, which is inherently a difficult problem as DNNs easily overfit noisy labels.

\noindent\textbf{Methods with robust loss functions.} There are numerous loss functions proposed to alleviate the influence of label noise. These include symmetric loss~\cite{ghosh2017robust}, variants of cross-entropy (CE) loss (generalized CE loss~\cite{wang2019symmetric}, symmetric CE loss~\cite{wang2019symmetric}, Taylor CE loss~\cite{feng2020can}), bootstrap loss~\cite{arazo2019unsupervised, reed2014training}, bilevel-optimization-based loss~\cite{jenni2018deep}, information-theoretic loss~\cite{xu2019l_dmi}, SIGUA loss~\cite{han2020sigua}. See also \cite{ma2020normalized} for a general framework for combining loss functions to enhance robustness to label noise.

\noindent\textbf{Noise model estimation methods.}
Label noise is modeled either explicitly by a noise transition matrix \cite{hendrycks2018using, patrini2017making}; or implicitly via graphical models \cite{xiao2015learning}, knowledge graphs \cite{li2017learning}, conditional random fields \cite{Vahdat2017toward}, residual nets \cite{hu2019weakly}, or joint embedding networks \cite{lee2018cleannet}. 
Once an underlying noise model has been well-estimated, the true labels can then be inferred with high confidence. 
For example, a good estimation of the noise transition matrix can be achieved by adding an extra noise model over the base model and simultaneously training both models \cite{jindal2016learning, sukhbaatar2014training}, 
albeit with certain strong assumptions.

\subsection{Preliminaries}
\label{preliminaries}
\textbf{Knowledge distillation} was first proposed by Hinton \textit{et al.} \cite{hinton2015distilling} for model compression,  
whereby the knowledge learned by a large teacher model $f_t(\cdot)$, is transferred to another smaller student model $f_s(\cdot)$, by applying a weighted average of two objective functions as follows:
\vspace*{-0.1em}
\begin{equation}
\label{distillation_loss_function}
\mathcal{L}(y,f_{s}(x))=\lambda l(y, f_{s}(x)) + (1-\lambda) l(f_{t}(x), f_{s}(x)),
\end{equation}
\vspace*{-1.1em}

\noindent where $l(\cdot)$ is the traditional loss function, and $\lambda$ is a parameter to balance the effect of the given labels and the outputs of the teacher model.

Assuming that a subset $\mathcal{D}_c$ of a given dataset is known to have correct labels, Li \textit{et al.} \cite{li2017learning} proposed a distillation-based method, where the teacher model $f_t(\cdot)$ is trained on $\mathcal{D}_c$ and the student model $f_s(\cdot)$ is subsequently trained using the loss function in \eqref{distillation_loss_function}. 
Hence, the student model would have a higher accuracy compared to the same model trained directly on the noisy dataset, as long as the error rate of the teacher model is less than the noise rate of the dataset.

A key merit of this distillation-based method is that it can be flexibly built on top of any algorithm for training the teacher and student models.
However, their method has two drawbacks: (i) The noisy labels remain invariant in the objective function, which affects the performance of the student model;
(ii) It requires a large clean set (e.g. the values $\frac{\vert \mathcal{D}_c \vert}{\vert \mathcal{D} \vert}=1$ and $\frac{1}{4}$ were used in their experiments) to train a sufficiently accurate teacher model. 

\textbf{PU learning} is a special type of semi-supervised learning that involves training a binary classifier on a set of positive data $\mathbf{P}$ and a set of unlabeled data $\mathbf{U}$~\cite{li2005learning}. 
Existing PU learning algorithms are classified into four approaches: (i) bias-based, (ii) heuristic-based, (iii) bagging-based, and (iv) risk-based. 

\begin{itemize}
\item A bias-based approach treats $\mathbf{U}$ as learnable weighted combinations of ``positive" and ``negative" classes \cite{elkan2008learning}.

\item A heuristic-based approach iteratively selects reliable negative samples from the unlabeled data via a two-step algorithm: (i) identify new reliable negative samples $\mathbf{N}$ from $\mathbf{U}$; (ii) train a binary classifier on $\mathbf{P}$ and $\mathbf{N}$ \cite{liu2002partially}.

\item A bagging-based approach uses an ensemble of classifiers trained on bootstrap subsets \cite{mordelet2014bagging}.

\item A risk-based approach estimates the misclassification risk by replacing the risk of negative samples with risk in terms of $\mathbf{P}$ and $\mathbf{U}$ given class prior \cite{du2014analysis}.
\end{itemize}

\textbf{Mixup} \cite{zhang2017mixup} is a data augmentation method proposed to favor linear relations on the samples, which involves augmenting the training dataset with convex combinations of pairs of examples and their labels. Specifically, given a pair of samples, ($x_i$, $y_i$) and ($x_j$, $y_j$), the augmented sample is generated via
\vspace*{-0.1em}
\begin{equation*}
\tilde{x}=\beta x_i + (1-\beta)x_j, \quad
\tilde{y}=\beta y_i + (1-\beta)y_j,
\end{equation*}
\vspace*{-1.1em}

\noindent where $\beta \sim \text{Beta}(\mu, \mu)$ follows a beta distribution for some parameter $\mu \in (0, \infty)$. We used $\mu=2$ in our experiments.

\textbf{Entropy regularization} \cite{tanaka2018joint} is introduced to concentrate the distribution of each prediction vector to one peak:
\begin{equation}
    \mathcal{L}_e=-\frac{1}{n}\sum_{i=1}^n\sum_{j=1}^C p_i(x_i)\log(p_i(x_i)),
\end{equation}
where $n$ is the number of instances, $C$ is the number of classes, and $p_i(\cdot)$ is the $i$-th entry of the output vector.
\section{Proposed Framework}
\label{sec: proposed method}

Consider a $C$-class noisy dataset with a small portion of clean set known to have correct labels. We denote the dataset by $\mathcal{D}=\mathcal{D}_c \cup \mathcal{D}_n$, where $\mathcal{D}_c$ and $\mathcal{D}_n$ represent this small clean set and the remaining noisy set, respectively.

\begin{algorithm}[H]
\caption{Proposed Method}\label{alg:algorithm}
\textbf{Inputs}: Number of classes $C$, clean set $\textstyle\mathcal{D}_c=\bigcup_{i= 1}^C\mathcal{D}_c^{(i)}$, noisy set $\mathcal{D}_n$, number of iterations $K$, number of ensemble models $N$, positive threshold $\alpha$, reliability criterion $\theta$, number of teacher models $N_t$. \\
\textbf{Intermediate teacher models}: $f^{(i)}_n$ ($1\leq n\leq N, 1\leq i\leq C$).\\
\textbf{Output}: Student classifier $f_s$.\\[-0.9em]
\begin{algorithmic}[1] 
\STATEx{// Clean data augmentation}
\FOR{$i=1$ \textbf{to} $C$}
  \STATE {$\hat{\mathcal{P}}^{(i)} \gets \emptyset$.}
  \FOR{$k=1$ \textbf{to} $K$} 
    \STATE{$m_k^{(i)} \gets \min\{\vert \mathcal{D}_c^{(i)} \vert, \vert \hat{\mathcal{P}}^{(i)} \vert\}$;  $m'^{(i)}_k \gets \frac{m_k^{(i)}+\vert \mathcal{D}_c^{(i)}\vert}{2}$.}
    \FOR{$n=1$ \TO $N$} 
      \STATE{\begin{varwidth}[t]{\linewidth}
      Randomly sample 
      $\mathcal{P}'$, $\mathcal{N}'$, and $\mathcal{N}''$, such that \par \hskip\algorithmicindent $\vert \mathcal{P}'\vert=m_k^{(i)}$, and $\vert \mathcal{N}'\vert=\vert \mathcal{N}''\vert=m'^{(i)}_k$.
      \end{varwidth}}
      \STATE{\begin{varwidth}[t]{\linewidth}
      Train $f^{(i)}_n$ on 
      $\Big\{\begin{smallmatrix}\mathcal{D}_c^{(i)}\cup \mathcal{P}' \text{ (positives)};\\ \mathcal{N}' \cup \mathcal{N}'' \text{ (negatives)}. \end{smallmatrix}$
      \end{varwidth}}
    \ENDFOR
    \STATE{$\hat{\mathcal{P}}^{(i)} \gets \left\{x\in\mathcal{D}_n \,\middle\vert\,  \vert \{n \mid f^{(i)}_n(x)\geq\alpha \} \vert \geq \theta \right\}$.}
    \STATE{Update $\mathcal{P}^{(i)}=\hat{\mathcal{P}}^{(i)}\cup\mathcal{D}_c^{(i)}$.}
  \ENDFOR
\ENDFOR
\STATE{$\mathcal{P} \gets \bigcup\limits_{i= 1}^C\mathcal{P}^{(i)}$.}
\STATEx{//Knowledge distillation}
\FOR{$n=1$ \textbf{to} $N_t$} 
      \STATE{\begin{varwidth}[t]{\linewidth}
      Train $f_t^{(n)}$ on a balanced bootstrap subset of $\mathcal{P}$.
      \end{varwidth}}
\ENDFOR
\STATE{Generate pseudo-labels for samples in $\mathcal{D}_n$ with teacher models}.
\STATE{Train a student model $f_s$ on $\mathcal{D}_c\cup\mathcal{D}_n$ with pseudo-labels.}
\RETURN {$f_s$}
\end{algorithmic}
\end{algorithm}

Our framework comprises two components, clean data augmentation and knowledge distillation.
In the first component, we introduce a new method of PU learning to train a filter on the clean set and generate the augmented clean set with the filter.
The filter and the clean set are updated iteratively, and the final augmented clean set is used to correct $\mathcal{D}_n$.
In the second component, we apply a variant of knowledge distillation, where the teacher model is trained on the augmented clean set, and the student model is trained on the entire dataset. (See Algorithm \ref{alg:algorithm} for a summary.)

\subsection{Clean Data Augmentation Component}
\label{clean data augmentation}

We first propose a tiered PU learning method to augment the small clean set with a two-step iterative method:
(i) Train an ensemble filter on the clean data with the idea of bagging \cite{Breiman1996}, which generates an ensemble of models separately trained on bootstrap subsets of the whole dataset;
(ii) Use the filter to choose reliable samples from the noisy set and update the clean set.
By repeatedly alternating between these two steps, we will gradually improve the filter and enlarge the clean set.

In contrast to existing PU learning approaches, which fix the given positive set throughout training, we update the positive set iteratively.
Also, despite having a similar two-step strategy, our approach is not heuristic-based: We do not need an initial distinguished set of reliable negative examples.

Let $\mathcal{P}=\hat{\mathcal{P}}\cup\mathcal{D}_c$ (resp. $\mathcal{P}^{(i)}=\hat{\mathcal{P}}^{(i)}\cup\mathcal{D}_c^{(i)}$) be the augmented clean set (for class $i$), where $\hat{\mathcal{P}}$ (resp. $\hat{\mathcal{P}}^{(i)}$) represents the additional reliable samples filtered from the noisy set (for class $i$).
Augmentation is done separately for each class, so for ease of explanation, consider a single class $i$.
Initialize $\hat{\mathcal{P}}^{(i)}=\emptyset$.
Next, run $K$ iterations, where in each iteration, form an ensemble filter consisting of $N$ binary classifiers $\{f^{(i)}_n\}_{n=1}^N$ to update the augmented clean dataset $\mathcal{P}^{(i)}$, described as follows: 
\begin{itemize}

\item \textbf{Train an ensemble filter $\{f^{(i)}_n\}_{n=1}^N$.}

The main idea is to separately train each binary classifier on a bootstrap subset of ``positive" data combined with a bootstrap subset of ``negative" data.
Note that $\mathcal{D}^{(i)}_c$ and $\textstyle\bigcup_{j\neq i}\mathcal{D}_c^{(j)}$ are already known to be absolutely positive samples and negative samples, respectively.
We shall obtain more positive (resp. negative) samples with a lower confidence level from the additional augmented clean set $\hat{\mathcal{P}}^{(i)}$ (resp. the remaining noisy set $\mathcal{D}_n\setminus \hat{\mathcal{P}}^{(i)}$).

Let $m_k^{(i)} := \min\{\vert \mathcal{D}_c^{(i)} \vert, \vert \hat{\mathcal{P}}^{(i)} \vert\}$ and $m'^{(i)}_k :=\frac{m_k^{(i)}+\vert \mathcal{D}_c^{(i)}\vert}{2}$ be two parameters to control the size of bootstrap subsets.
For each binary classifier $f^{(i)}_n$, we shall sample subsets $\mathcal{P}'\subseteq\hat{\mathcal{P}}^{(i)}$, $\mathcal{N}'\subseteq\mathcal{D}_n\setminus \hat{\mathcal{P}}$ and $\mathcal{N}''\subseteq \textstyle\bigcup_{j\neq i}\mathcal{D}_c^{(j)}$ uniformly at random, such that $\vert \mathcal{P}'\vert=m_k^{(i)}$, $\vert \mathcal{N}'\vert=\vert \mathcal{N}''\vert=m'^{(i)}_k$.

Next, we train $f^{(i)}_n$ on $\mathcal{D}_c^{(i)}\cup \mathcal{P}'$ as the positives, and $\mathcal{N}' \cup \mathcal{N}''$ as the negatives.

Note that $f^{(i)}_1, \dots, f^{(i)}_N$ are trained on different random subsets, each with an equal number of positive and negative samples. 
Hence, the aggregation of these $N$ models would have lower bias.

\item \textbf{Generate the augmented clean set $\mathcal{P}^{(i)}=\hat{\mathcal{P}}^{(i)}\cup\mathcal{D}_c^{(i)}$.}

Using the ensemble filter generated in the previous step, we select samples from the noisy set that are identified as positive samples with relatively high confidence. 
Specifically, we define the set of additional positive samples. 
\begin{equation}
\hat{\mathcal{P}}^{(i)}=\left\{x\in\mathcal{D}_n
\,\middle\vert\, 
 \vert \{n \mid f^{(i)}_n(x)\geq\alpha \} \vert \geq \theta \right\},
\end{equation}
where the reliability criterion $\theta$ and confidence threshold $\alpha$ are two hyperparameters designed to control the confidence level of $\hat{\mathcal{P}}^{(i)}$. 
The set of positive samples for each model $f^{(i)}_n$ is given by $\{x \vert f^{(i)}_n(x)>\alpha\}$, thus $\hat{\mathcal{P}}^{(i)}$ is composed of samples $x$ for which there exist at least $\theta$ binary classifiers ($f^{(i)}_\ast$) that classify $x$ as positive data. 

To complete this iterative step, let $\mathcal{P}^{(i)} = \hat{\mathcal{P}}^{(i)} \cup \mathcal{D}_c^{(i)}$, and samples in $\mathcal{P}^{(i)}$ are automatically relabeled with label $i$.
\end{itemize}
Repeat these two steps for $K$ iterations. 

In general, we convert a multi-class semi-supervised problem into $C$ binary classification problems that are easier to solve. Take class $i$ for example, we have a small clean “positive” set $\mathcal{D}_c^(i)$ at the beginning, and we sample a “negative” bootstrap set from $\mathcal{D}_n$ to train each binary model, where roughly $\frac{C-1}{C}\times 100\%$ samples are true negatives. Intuitively, the negative set sampled in this way is more reliable than other semi-supervised methods that are based on similarity measures, and is easier to implement.

\subsection{Knowledge Distillation Component}
Given the augmented clean dataset $\mathcal{P}$, we apply a variant of knowledge distillation to train a teacher-student model.
In contrast to the original distillation method \cite{hinton2015distilling, li2017learning}, we instead use a casewise parameter $\lambda(x_j)$ based on the maximum entry of the prediction vector of teacher model.

We first train the teacher model on the augmented clean set $\mathcal{P}$ with reassigned labels. Next, we train the student model on the entire dataset $\mathcal{D}$. (The given labels are already corrected at the end of the previous component.)
If $\mathcal{P}$ is class-balanced, we directly train a multi-classifier on $\mathcal{P}$ as the teacher model. 
Otherwise, we train an ensemble of several (e.g. 5) classifiers on balanced bootstrap subsets of $\mathcal{P}$, where each bootstrap subset contains $\smash{\underset{i}{\operatorname{min}} \vert \mathcal{P}^{(i)} \vert}$ samples for every class.

Note that the loss function given in \eqref{distillation_loss_function} is equivalent to a normal loss function (e.g. cross-entropy) applied to pseudo-label $\hat{y}$, which is generated as a weighted combination of the teacher model output and the (corrected) given label. There are three ways to generate the pseudo-labels:

\begin{itemize}
\item Soft bootstrap label:
\vspace*{-0.4em}
\begin{equation}
\label{soft bootstrap label formulation}
\hat{y}_j=\lambda(x_j)f_t(x_j)+(1-\lambda(x_j))y_j,
\end{equation}
\vspace*{-1.5em}

\noindent where $f_t(x_j)$ is the average prediction vector of 5 teacher models, $y_j$ is the one-hot vector of the $j$-th corrected given label, and $\lambda(\cdot)$ is a function to measure the confidence level of the teacher models for $x_j$, defined by
\vspace*{-0.7em}
\begin{equation}
\label{casewise parameter}
\lambda(x_j) =
\begin{cases}
\lambda, &\mbox{if } \max(f_t(x_j))\geq\eta; \\
0, & \mbox{if } \eta > \max(f_t(x_j));
\end{cases}
\end{equation}
\vspace*{-0.9em}

\noindent where $\eta$ is used to control the confidence level of the pseudo-labels, and $\lambda$ is used to keep the balance between the prediction of teacher model and given label. 

\item Hard bootstrap label: The definition is similar to soft bootstrap label, but the prediction vector $f_t(x_j)$ in \eqref{soft bootstrap label formulation} is replaced by a one-hot vector, where the entry of value $1$ is the maximum entry in $f_t(x_j)$, i.e. $\operatorname{argmax} f_t(x_j)$.

\item Hard label: 
\vspace*{-0.1em}
\begin{equation}
\label{hard label formulation}
\hat{y}_j =
\begin{cases}
\operatorname{argmax} \ \ f_t(x_j), &\mbox{if } \max(f_t(x_j))\geq\eta; \\
\operatorname{argmax} \ \ y_j, & \mbox{if } \eta > \max(f_t(x_j)).
\end{cases}
\end{equation}
\end{itemize}

\section{Experiments}

\label{experiment}
\subsection{Experiment Setup}
\subsubsection{Datasets}
\label{dataset}
\textbf{CIFAR-10}.
The CIFAR-10 dataset \cite{krizhevsky2009learning} contains  50,000 training images and 10,000 test images in 10 classes, with 6,000 images per class. Each image has a size of $32\times 32\times 3$. 
Let the ratio of the original clean set $\mathcal{D}_c$ and the noise level of the entire dataset $\mathcal{D}$ be denoted by $\pi$\% and $r$\% respectively. 
Within each class, we sampled $\frac{\pi\%}{10}\times50,000$ images uniformly at random from the training set to form the original clean data $\mathcal{D}_c$.  
Then, we added two types of synthetic label noise to the remaining $(100-\pi)$\% data as follows:  
\begin{itemize}
\item \textbf{Symmetric noise}. We randomly chose $\frac{100r}{100-\pi}\%$ samples from the remaining $(100-\pi)\%$ data, and for each chosen sample, we randomly assigned a new label.
We require a $\frac{100}{100-\pi}$ factor to take into consideration the original $\pi$\% clean set used in our method;
\item \textbf{Asymmetric noise}. We followed the definition in \cite{patrini2017making}, where label noise for pairs of semantically similar classes (CAT$\leftrightarrow$DOG,  DEER$\rightarrow$HORSE, BIRD$\rightarrow$AIRPLANE, TRUCK$\rightarrow$AUTOMOBILE) was generated by randomly assigning $\frac{100r}{100-\pi}\%$ samples from each objective class with the target label. Note that the noise level defined in \cite{patrini2017making} refers the class noise level. 
As we have $5$ objective classes, the overall noise level of the dataset is $0.5r\%$.
\end{itemize} 

\textbf{Clothing1M}.
The clothing1M dataset \cite{xiao2015learning} is a real-world image dataset with both noisy and clean labels. 
There are over a million clothing images in 14 classes collected from online shopping websites, and a noisy label is automatically assigned to each image based on the keywords in surrounding text. 
A manually labeled clean subset with 72,409 images is provided.

\subsubsection{Baselines}
For CIFAR-10, we compared our framework with the following baselines, using the code provided in the respective papers. 
For those methods that add synthetic noise via a transition matrix, we multiply the noise level by factor 0.9 $(i.e., \frac{N_{class}-1}{N_{class}})$ for fair comparison. 
\begin{itemize}
    \item Mixup\cite{zhang2017mixup}, which alleviates the effect of noisy labels by training on convex combinations of pairs of samples.
    \item Joint Optimization \cite{tanaka2018joint}, which alternately updates network parameters and labels during training.
    \item Co-teaching\cite{han2018co}, which simultaneously trains two neural networks and cross-updates the network parameters. 
    \item Loss Correction\cite{arazo2019unsupervised}, which estimates wrong label probabilities and corrects the loss with a beta mixture model. 
    \item JoCoR\cite{wei2020combating}, which jointly trains two networks using a joint loss with co-regularization.
    \item DivideMix\cite{li2020dividemix}, which divides the dataset into two subsets, and concurrently trains two networks in a semi-supervised manner using MixMatch.
\end{itemize}

\subsubsection{Implementation Details}
For CIFAR-10, we used a Pre-Activation ResNet-18 \cite{he2016identity} and an SGD optimizer with a momentum of 0.9, a weight decay of $10^{-4}$, and batch size of 128. For preprocessing, the images were normalized, and augmented by random horizontal flipping and random 32$\times$32 cropping with padding=4. In the clean set augmentation step, each model was trained over 30 epochs. The learning rate was initialized at 0.01 and was divided by 10 after 20 epochs. The teacher models and the student model were trained over 100 epochs each. The learning rate was initialized at 0.05 and was divided by 10 after 30, 50, and 80 epochs.

For Clothing1M, we follow the configuration used in previous works \cite{arazo2019unsupervised,li2019learning,tanaka2018joint, xu2019l_dmi, zhang2019metacleaner, lee2018cleannet, han2019deep}, i.e. we used a ResNet-50 \cite{he2016identity} pre-trained on ImageNet. We used an SGD optimizer with a momentum of 0.9, a weight decay of $10^{-4}$, a cross-entropy loss function, and a batch size of 32.
For preprocessing, the images were normalized, and augmented by random horizontal flipping and centered 224$\times$224 cropping. 
In the clean set augmentation step, each model was trained over 20 epochs. The learning rate was initialized at 0.03 and was divided by 10 after 10 and 15 epochs. We trained the student model and each teacher model for 5 epochs. The learning rate was initialized at 0.001 and was divided by 10 after 3 epochs.

Code is available at \url{https://github.com/Xu-Jingyi/PUDistill}.

\begin{table}[tb]
\caption{The precision and size of the augmented clean set for CIFAR-10 (test set) and Clothing1M (validation set).}
\label{Table of augmented clean set}
\centering
\begin{tabular}{lrr|rr}
\toprule
\multirow{2}{*}{Class} & \multicolumn{2}{c|}{CIFAR10}   & \multicolumn{2}{c}{Clothing1M} \\ \cmidrule{2-5}
 & \multicolumn{1}{r}{Precision} & \multicolumn{1}{r|}{Size} & \multicolumn{1}{r}{Precision} & \multicolumn{1}{r}{Size} \\ \midrule
0   & 0.92   & 413   & 0.94   & 298   \\
1   & 0.98   & 586   & 0.83   & 142   \\
2   & 0.93   & 281   & 0.46   & 166   \\
3   & 0.80   & 219   & 0.95   & 736   \\
4   & 0.90   & 306   & 0.78   & 651   \\
5   & 0.93   & 368   & 0.93   & 554   \\
6   & 0.96   & 462   & 0.69   & 247   \\
7   & 0.97   & 456   & 0.54   & 273   \\
8   & 0.96   & 575   & 0.93   & 498   \\
9   & 0.97   & 538   & 0.95   & 480   \\
10  & - &  - & 0.96   & 272   \\
11  & - &  - & 0.58   & 173   \\
12  & - &  - & 0.82   & 680   \\
13  & - &  - & 0.91   & 624  \\ \midrule
\multirow{2}{*}{Overall} & \multirow{2}{*}{0.94} & 4,204 
& \multirow{2}{*}{0.85} & 5,794  \\
 & & (out of 10,000) & & (out of 14,313) \\
\bottomrule
\end{tabular}
\end{table}
\begin{table}[htb!]
\caption{Accuracy of the ensemble teacher model, which is evaluated on the subset  $\{x\in \mathcal{D}_{test}\mid\max(f_t(x))\geq\eta\}$ of the test set (resp. evaluated on the validation set) for CIFAR-10 (resp. Clothing1M) with various threshold values $\eta$.\\[-1.5em]}
\label{teacher model}
\centering
\begin{tabular}{lrr|rr}
\toprule
\multirow{2}{*}{$\eta$} & \multicolumn{2}{c|}{CIFAR10} & \multicolumn{2}{c}{Clothing1M} \\ \cmidrule{2-5}
 & Accuracy (\%) & Size   & Accuracy (\%)   & Size \\ \midrule
0.9 & 99.3   & 1,313  & 94.6  & 5,635   \\
0.8 & 98.0   & 4,845  & 91.3  & 8,219   \\
0.7 & 95.6   & 6,492  & 88.1  & 9,950   \\
0.6 & 93.0   & 7,561  & 85.1  & 11,407  \\
0.5 & 89.8   & 8,487  & 82.0  & 12,808  \\
0.0 & 82.5   & 10,000 & 77.7  & 14,313  \\ \bottomrule
\end{tabular}
\end{table}

\subsection{Comparison between Different Ratios of Clean Set}
\begin{figure}[tb]
\centering
\includegraphics[width=80mm]{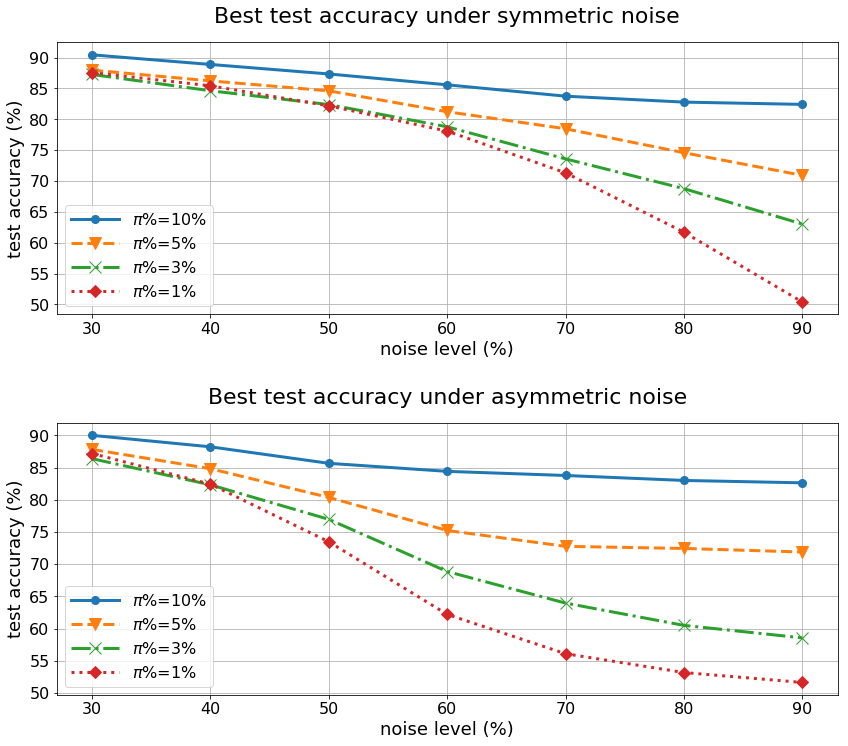}
\caption{Comparison of the best test accuracies (average of 5 trials) corresponding to different clean ratios $(\pi\%)$ on  CIFAR-10 with symmetric noise (upper) and asymmetric noise (lower). All experiments were implemented using soft-bootstrap pseudo-labels.}
\label{Comparison clean ratio}
\end{figure}

We applied bootstrapping when training the teacher model on the augmented clean set to (i) mitigate the problem of imbalance, and (ii) reduce the training time. Specificically, we sampled a balanced subset of the augmented clean set in each epoch, and trained the teacher model on the bootstrap subset, instead of the entire augmented clean set. In order to improve robustness, we also used mixup data augmentation and entropy regularization (cf. Section \ref{preliminaries}) to train the teacher models and student model. The detailed ablation study and elaboration of the effectiveness of each technique is given in Section \ref{ablation study}.

\begin{figure}[htb!]
\centering
\includegraphics[width=80mm]{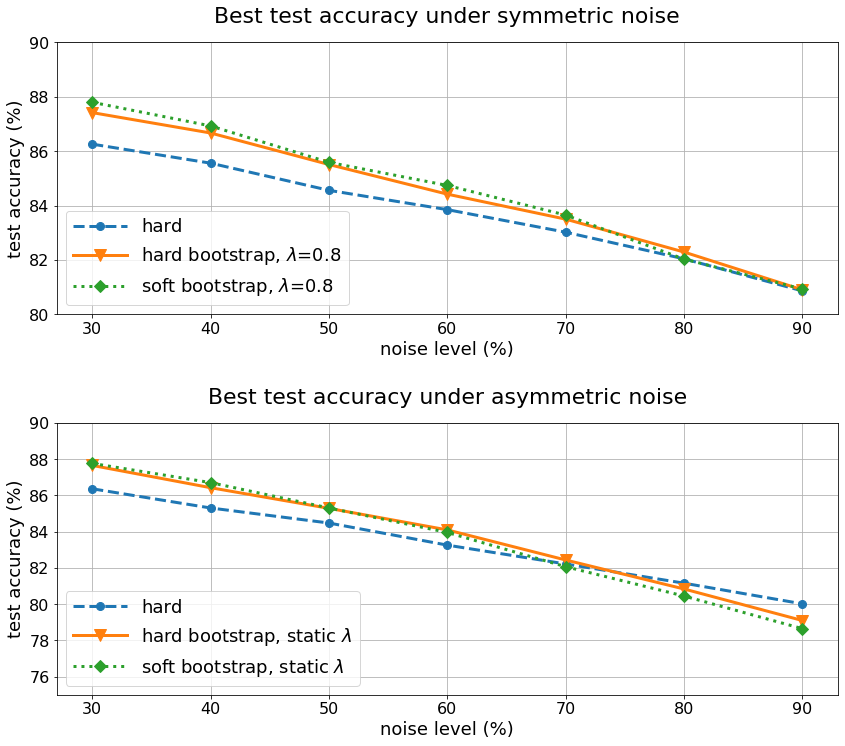}
\caption{Comparison of the best test accuracies (average of 5 trials) corresponding to different types of pseudo-labels on CIFAR-10 with symmetric noise (upper) and asymmetric noise (lower). We fixed $\eta=0.7$ and $\pi\%=10\%$ in the experiments.}
\label{Comparison types of label}
\end{figure}

\subsection{Evaluation of Augmented Clean Set}
To show the effectiveness of our clean data augmentation algorithm, we reported the precision and number of augmented clean samples in TABLE \ref{Table of augmented clean set}, where the results were evaluated on test set for CIFAR-10 and validation set for Clothing1M. As described in Section \ref{clean data augmentation}, we introduced a threshold $\alpha$ to filter ``positive" samples, and a hyperparameter $\theta$ to control the size of augmented clean set. In our experiment, we fixed $\alpha=0.9$, $\theta=19$ (out of 20) for CIFAR-10, and $\alpha=0.95$, $\theta=10$ (out of 10) for Clothing1M. Our method extracted more than 40\% samples while achieved overall precision 0.94 and 0.85 for CIFAR-10 and Clothing1M, respectively. 

\subsection{Evaluation of Teacher Model}
As given in \eqref{soft bootstrap label formulation}--\eqref{hard label formulation}, 
the teacher model is used to generate pseudo-labels for the training of the student model.
In order to guarantee the accuracy of the pseudo-labels, we only modified the labels of samples in $\{x_j\in \mathcal{D}_n \mid\max(f_t(x_j))\geq\eta \}$, while the pseudo-labels of the remaining samples were kept as the given labels.
Here $f_t(\cdot)$ is the average output vector of 5 models, and $\eta$ is the threshold to control the confidence level of the ensemble teacher model.
We evaluated the ensemble teacher model 
for different values of $\eta$; see TABLE \ref{teacher model}.

\begin{table}[!tb]
\caption{Best test accuracies of different methods on the Clothing1M dataset. For all baselines, we use the reported results in the respective papers.\\[-1.5em]}
\label{clothing reuslt}
\centering
\begin{tabular}{llr}  
\toprule
\# & Method   &  Accuracy (\%) \\
\midrule
1 & Cross Entropy \cite{tanaka2018joint} &  68.94 \\ 
2 & Forward \cite{patrini2017making}  &  69.84 \\
3 & JoCoR \cite{wei2020combating}  & 70.30 \\
4 & Loss correction \cite{arazo2019unsupervised}     & 71.00\\
5 & Joint opt. \cite{tanaka2018joint}   &    72.23 \\
6 & $\mathcal{L}_{\text{DMI}}$ \cite{xu2019l_dmi}  & 72.46 \\
7 & Metacleaner \cite{zhang2019metacleaner}  &   72.50 \\
8 & Meta learning\cite{li2019learning}  &   73.47 \\

9 & DeepSelf \cite{han2019deep}  &   74.45 \\
10 & CleanNet \cite{lee2018cleannet}  &   74.69 \\

11 & DivideMix\cite{li2020dividemix}  & 74.76 \\
\midrule
12 & our method  & \textbf{77.70} \\
\bottomrule
\end{tabular}
\end{table}

\begin{table}[!tb]
\caption{Ablation study results in terms of teacher model's test accuracy (\%) on CIFAR-10 and Clothing1M.\\[-1.5em]}
\label{ablation teacher}
\centering
\begin{tabular}{lrr}
\toprule
Methods  & CIFAR-10 & Clothing1M \\\midrule
Standard & 76.88$\pm$1.82   & 77.27$\pm$0.11  \\\midrule
+ mixup  & 80.68$\pm$2.48   & 77.57$\pm$0.05  \\\midrule
+ bootstrap    & 77.31$\pm$1.16   & 76.97$\pm$0.47 \\\midrule
+ entropy reg. + bootstrap & 77.12$\pm$1.30   & 77.68$\pm$0.04  \\\midrule
+ mixup + bootstrap    & 81.05$\pm$1.70   & 77.75$\pm$0.10 \\\midrule
+ mixup + bootstrap + entropy reg. & 81.45$\pm$1.45   &  77.18$\pm$0.05  \\
\bottomrule
\end{tabular}
\end{table}

\subsection{Comparison between Different Types of Pseudo-labels}
To show the difference between three types of pseudo-labels introduced in \eqref{soft bootstrap label formulation}--\eqref{hard label formulation}, we compared the best test accuracies on CIFAR-10 with different levels of symmetric noise and asymmetric noise in Fig. \ref{Comparison types of label}.
For both symmetric and asymmetric noise, hard label performed worse than two bootstrap labels at low noise levels, and hard bootstrap label had a similar performance compared with soft bootstrap label.
Intuitively, hard label resulted in lower accuracies because it failed to take the given label into consideration for those samples with high confidence, i.e. $\max(f_t(x_j))\geq\eta$. However, this problem was mitigated by the case-wise pattern of pseudo-labels, where the pseudo-label was only assigned to those samples that teacher model was confident on, and the labels of the remaining samples remained the given labels. Hence this case-wise pattern can avoid introducing additional noise because of inaccurate teacher model, and it happened to shorten the gap between hard labels and the other two bootstrap labels.

As described previously, our method uses an initial small clean set $\mathcal{D}_c$, with proportion $\pi\%:=\frac{\vert \mathcal{D}_c \vert}{\vert \mathcal{D}_c\cup\mathcal{D}_n \vert}$. To study the effect of $\pi$ on the test accuracy, we applied our algorithm to CIFAR-10 with different noise levels for both symmetric and asymmetric noise. \mbox{Fig. \ref{Comparison clean ratio}} shows that the gap between curves of different clean ratios is not large under low noise levels, and our method performs well even with only a $1\%$ clean set. However, at high noise levels, the curves for low clean ratios dramatically dropped, especially for asymmetric noise, while the curve for $10\%$ clean ratio was relatively flat.

\subsection{Comparison with State-of-the-art}
\subsubsection{Experiments on CIFAR-10}
\begin{table*}[!tb]
\caption{Average (5 trials) and standard deviation of the best test accuracies  of different methods on the CIFAR-10 dataset with semantic asymmetric noise. The highest accuracy for each noise level is boldfaced.}
\label{cifar_asyn}
\centering
\begin{tabular}{lrrrrrrr}
\toprule
\multirow{2}{*}{Method} & \multicolumn{7}{c}{Noise Level (\%)} \\ \cmidrule{2-8} 
  & 30  & 40  & 50  & 60  & 70  & 80  & 90  \\ \midrule
Cross Entropy   & 88.30$\pm$0.42 & 84.28$\pm$0.65 & 77.40$\pm$1.52 & 67.54$\pm$1.21 & 61.72$\pm$0.30 & 57.24$\pm$0.31 & 52.70$\pm$0.21   \\ \midrule
Mixup\cite{zhang2017mixup}  & 90.53$\pm$0.70 & 86.59$\pm$1.13 & 78.67$\pm$0.93 & 69.19$\pm$1.16 & 62.52$\pm$1.32 & 57.90$\pm$2.1 & 53.30$\pm$0.80    \\ \midrule
Joint Optimization\cite{tanaka2018joint}  & 92.01$\pm$0.21 & \textbf{89.56$\pm$0.78} & 84.56$\pm$0.94 & 78.21$\pm$0.32 & 76.70$\pm$0.11 & 76.44$\pm$0.21 & 76.00$\pm$0.14  \\ \midrule
Co-teaching\cite{han2018co}  & 84.50$\pm$0.41 &	70.69$\pm$3.53 &	54.29$\pm$1.30 &	48.76$\pm$0.92 &	46.40$\pm$0.89 &	45.66$\pm$1.77 &	44.39$\pm$1.53 \\\midrule
Loss Correction\cite{arazo2019unsupervised}  & 90.87$\pm$0.23 & 87.21$\pm$0.22 & 74.63$\pm$1.08 & 58.82$\pm$1.08 & 58.06$\pm$0.09 & 53.72$\pm$3.32 & 53.04$\pm$3.59    \\ \midrule
JoCoR\cite{wei2020combating}  &76.57$\pm$2.67 &	67.74$\pm$4.23 	& 56.54$\pm$1.22 &	48.52$\pm$1.84 &	46.22$\pm$0.80 & 43.74$\pm$0.81 & 43.06$\pm$1.89 \\ \midrule
DivideMix\cite{li2020dividemix}  & \textbf{93.95$\pm$0.06} & 84.43$\pm$0.94 & 73.73$\pm$1.07 & 60.13$\pm$2.40 & 53.18$\pm$3.99 & 50.60$\pm$0.546 & 49.42$\pm$0.33 \\ \midrule
Our Method &  90.00$\pm$0.22 & 88.22$\pm$0.26 & \textbf{85.98$\pm$0.45} & \textbf{84.41$\pm$0.15} & \textbf{83.77$\pm$0.12} & \textbf{83.00$\pm$0.10} & \textbf{82.63$\pm$0.25} \\ 
 \bottomrule
\end{tabular}
\end{table*}

\begin{figure*}[tb]
\centering
\includegraphics[width=180mm]{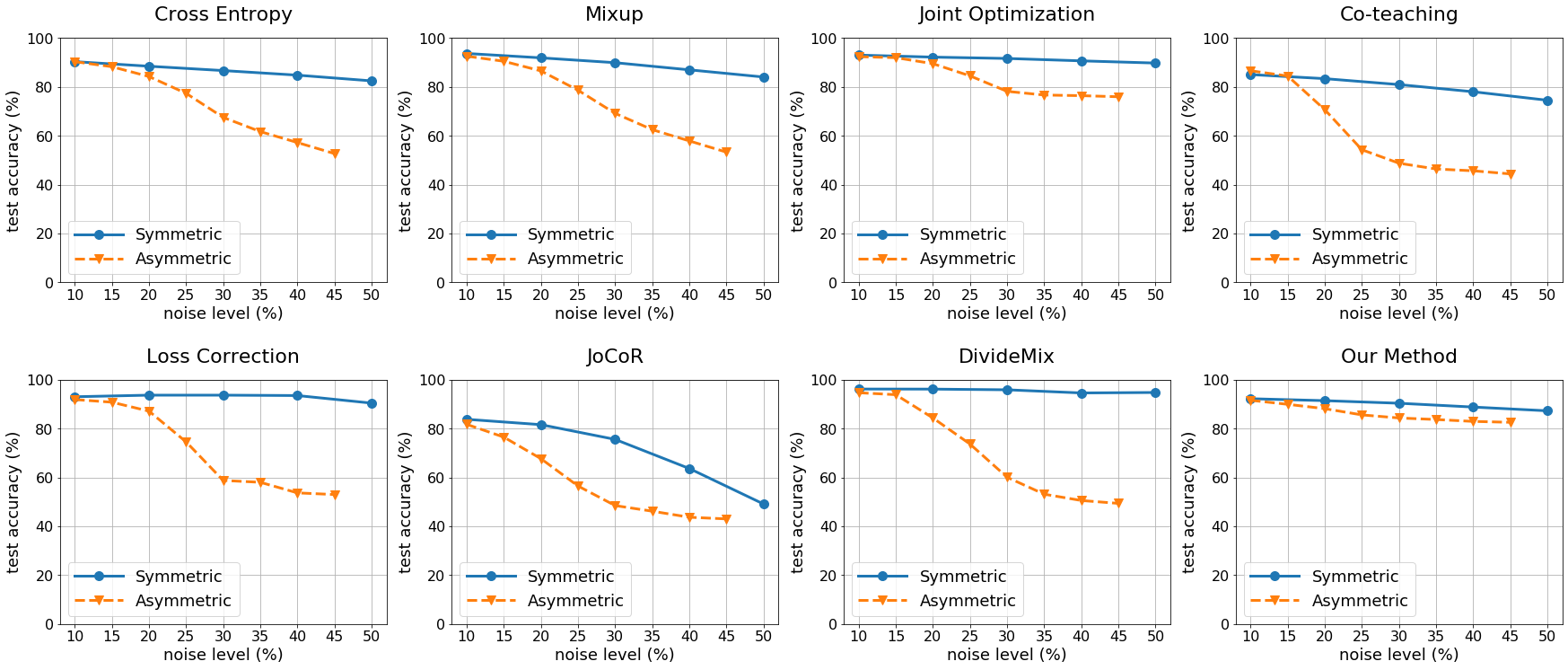}
\caption{Comparison of the performance of different methods under symmetric noise and asymmetric noise. The noise level here refers to overall noise level within the dataset, e.g. $r\%$ asymmetric noise would correspond to $0.5r\%$ overall noise level. 
}
\label{cifar_fig}
\end{figure*}

We reproduced the results of the baselines \cite{tanaka2018joint,arazo2019unsupervised,zhang2017mixup,han2018co,wei2020combating, li2020dividemix} using the same configuration that we used to evaluate our proposed method: 
(i) We used the same architecture, Pre-Activation ResNet-18 \cite{he2016identity}; 
(ii) We multiplied the noise level in \cite{han2018co, wei2020combating} by a factor 0.9, and multiplied the noise level in our method with a factor $\frac{100}{100-\pi}$ to keep the expected proportion of incorrect labels equal across all methods. 
The comparison results between our method and all the baselines on the CIFAR-10 dataset with asymmetric noise are reported in TABLE \ref{cifar_asyn}.

To evaluate the robustness of each method against different noise models, we compared the accuracies of each mothod with different noise levels for symmetric and asymmetric noise in \mbox{Fig. \ref{cifar_fig}}. As claimed in Section \ref{dataset}, the noise level of asymmetric defined in \cite{patrini2017making} is the class noise level within the corrupted classes, hence the overall noise level of the dataset is actually $0.5r\%$. In order to compare the performance of each method between symmetric model and asymmetric model, we need to modify the noise level of asymmetric noise by a factor $0.5$. The curves corresponding to symmetric and asymmetric noise model of our method were very close and flat across all of the noise levels. However, the gap between two lines for other 7 baselines became larger as the noise level increased, and the curves of some methods dramatically dropped, especially under asymmetric case. Our method is robust not only over all noise levels but also against different noise models because of the effectiveness of the clean data augmentation step. By treating all the noisy samples as unlabeled samples and applying a tired PU learning method, our augmentation process is totally independent of the given noisy labels, thereby explaining this robustness. 

\subsubsection{Experiments on Clothing1M}

We directly compared our experiment result with accuracies reported in papers that used the same model architecture, i.e. a ResNet-50 \cite{he2016identity} pre-trained on ImageNet; see TABLE \ref{clothing reuslt}. 
We trained the student model over 5 epochs and achieved a highest test accuracy 77.70\%; this is the highest accuracy among all methods in TABLE \ref{clothing reuslt}.

\subsection{Ablation Study}
\label{ablation study}
To study the effect of mixup data augmentation, bootstrapping, and entropy regularization on our method, we conducted an ablation study on the teacher model (see TABLE \ref{ablation teacher}) and student model (see TABLE \ref{ablation student}), where the ``standard" method means our method without any of these three techniques. Below, we consolidate some insights obtained.

\begin{itemize}
\item All three techniques helped to improve the accuracies slightly. Mixup data augmentation has the largest effect.
\item The effect of mixup is more significant under symmetric noise than under asymmetric noise. 
Intuitively, mixup improves the robustness to label noise because the combination of a pair of samples might partially correct the wrong labels when one sample accidentally contains the true label of the other noisy sample. For asymmetric noise, the label flips occur only between similar classes, so ``accidental label correction" is less likely.
\end{itemize}

\section{Conclusion and Further Remarks}
Our proposed framework is robust to all noise levels. This robustness is universal since the performance at any given noise level is not sensitive to variations in the noise model. 
We achieved state-of-the-art accuracy on Clothing1M, a dataset with real-world label noise, and our experiments on CIFAR-10 with asymmetric semantic label noise show superior outperformance over all baselines. 
Crucially, we achieve universal robustness because our clean data augmentation process does not use the labels of noisy samples; we only require a small clean subset. 
Our framework can be built on top of any classification algorithm (not necessarily using DNNs). Therefore, our framework is versatile, robust, and widely applicable to any classification tasks for datasets with arbitrary label noise.

\begin{table*}[!tb]
\caption{Ablation study results in terms of student model's test accuracy (\%) on CIFAR-10. We set the ratio of original clean set=0.1, $\eta=0.9$, and $\lambda=0.5$.\\[-1.5em]}

\label{ablation student}
\centering
\begin{tabular}{lrrrrrrrr}
\toprule
Noise type           & \multicolumn{4}{c}{Symmetric} & \multicolumn{4}{c}{Asymmetric} \\ \midrule
Method/Noise level   & 30\%  & 50\%  & 70\%  & 90\%  & 30\%   & 50\%  & 70\%  & 90\%  \\\midrule
Standard & 84.74$\pm$0.32 &  81.06$\pm$0.33 & 77.39$\pm$0.37 & 73.29$\pm$0.66  & 87.32$\pm$0.47 & 82.91$\pm$0.46 & 76.34$\pm$0.84 & 70.32$\pm$1.76  \\ \midrule
+Mixup & 88.90$\pm$0.23 &  85.34$\pm$0.34 & 81.82$\pm$0.19 & 76.94$\pm$0.45 & 89.30$\pm$0.27 & 84.93$\pm$0.32 & 77.72$\pm$1.50 & 71.25$\pm$2.51 \\\midrule
+Entropy reg. & 86.52$\pm$0.21 &  83.26$\pm$0.36 & 79.68$\pm$0.38 & 74.58$\pm$0.40 & 87.41$\pm$0.23 & 82.98$\pm$0.44 & 76.16$\pm$1.14 & 70.33$\pm$1.93  \\\midrule
+Mixup +Entropy reg. & 90.10$\pm$0.17 &  87.12$\pm$0.26 & 83.52$\pm$0.10 & 78.88$\pm$0.49 & 90.00$\pm$0.22 & 85.66$\pm$0.26 & 77.20$\pm$1.61 & 70.52$\pm$2.73 
 \\\midrule
\bottomrule
\end{tabular}
\end{table*}

\bibliographystyle{IEEEtran}

\bibliography{bibliography}

\end{document}